\documentclass[letterpaper, 10 pt, conference]{ieeeconf}  

\IEEEoverridecommandlockouts          

\usepackage{graphicx} 
\usepackage{amsmath} 
\usepackage{amssymb}  
\usepackage{booktabs}
\usepackage{tabularx}
\usepackage{makecell}
\usepackage[colorlinks=true, urlcolor=blue, citecolor=black, linkcolor=black]{hyperref}
\title{\LARGE \bf VLN-AVP: Zero-Shot Vision-Language Navigation with Hybrid Long-Short-Term Memory for Autonomous Valet Parking }

\author{Yijian Li$^{1}$, Xiangru Mu$^{1}$, Changze Li$^{1}$, Hantian Shi$^{1}$, Jiyuan Cai$^{2}$,\\
Jia Cai$^{2}$, Xiaoxue Liu$^{2}$, Yajing Sun$^{2}$, Ming Yang$^{1}$, and Tong Qin$^{1}$%
\thanks{$^{1}$Yijian Li, Xiangru Mu, Changze Li, Hantian Shi, Ming Yang,
and Tong Qin are with Shanghai Jiao Tong University, Shanghai, China.}%
\thanks{$^{2}$Jiyuan Cai, Jia Cai, Xiaoxue Liu, and Yajing Sun are with
Yinwang Intelligent Technology Co., Ltd.}%
\thanks{Tong Qin is the corresponding author: {\tt\small qintong@sjtu.edu.cn}.}%
\thanks{This work was supported by the National Natural Science Foundation of China (Grant No. 62403312).}%
}

\begin{document}
    \maketitle
    \thispagestyle{empty}
    \pagestyle{empty}

    \begin{abstract}
        Existing methods in Autonomous Valet Parking (AVP) typically rely on pre-built maps, which severely restricts their scalability to unseen environments and open-vocabulary targets. Inspired by the application of Vision-Language Models (VLMs) in Vision-Language Navigation (VLN) tasks, we propose VLN-AVP, a zero-shot navigation framework for AVP tasks. By combining the precise spatial perception of a Bird's-Eye-View (BEV) model with the general intelligence of VLMs, our framework 1) eliminates the dependency on pre-built maps, 2) interprets semantic environmental contexts in parking scenarios, and 3) enables intuitive navigation following natural language instructions. Specifically, we introduce a hybrid memory system: a short-term perception memory tracks semantic visual cues to address the limitations of VLM's single-frame reasoning in existing methods, while a long-term topological memory facilitates stable policy learning from past experiences. To bridge the gap in existing benchmarks, we also present the VLN-AVP dataset and benchmark. Featuring 10 high-fidelity parking scenes and over 1,000 navigation episodes, it has the largest number of garage scenes to date and is the first VLN benchmark for underground parking. Extensive experiments demonstrate that in simulation, our method achieves an over 25\% improvement in success rate compared to VLN methods and an over 15\% improvement compared to other autonomous driving methods. Furthermore, it attains a leading success rate in real-world vehicle experiments, proving its practical feasibility. 
    \end{abstract}

    \section{Introduction}

    As a critical component of intelligent transportation systems, Autonomous Valet
    Parking (AVP) integrates autonomous driving technologies such as perception,
    localization, and trajectory planning, significantly improving the efficiency and user experience of parking. Current AVP methodologies \cite{avp-structure, avp-slam} usually rely on pre-built maps, such as high-definition (HD) or vectorized maps \cite{avp-slam,liu2024less}. These maps require pre-collected environmental data to construct scene representations of parking garages before performing parking tasks in corresponding scenarios, which constrains the operational flexibility and environmental adaptability of AVP systems. In addition, due to the lack of human-machine interaction capabilities, current AVP systems are limited to performing predefined tasks, such as navigating to a designated parking slot.

    \begin{figure}[!t]
        \centering
        \includegraphics[width=.95\linewidth]{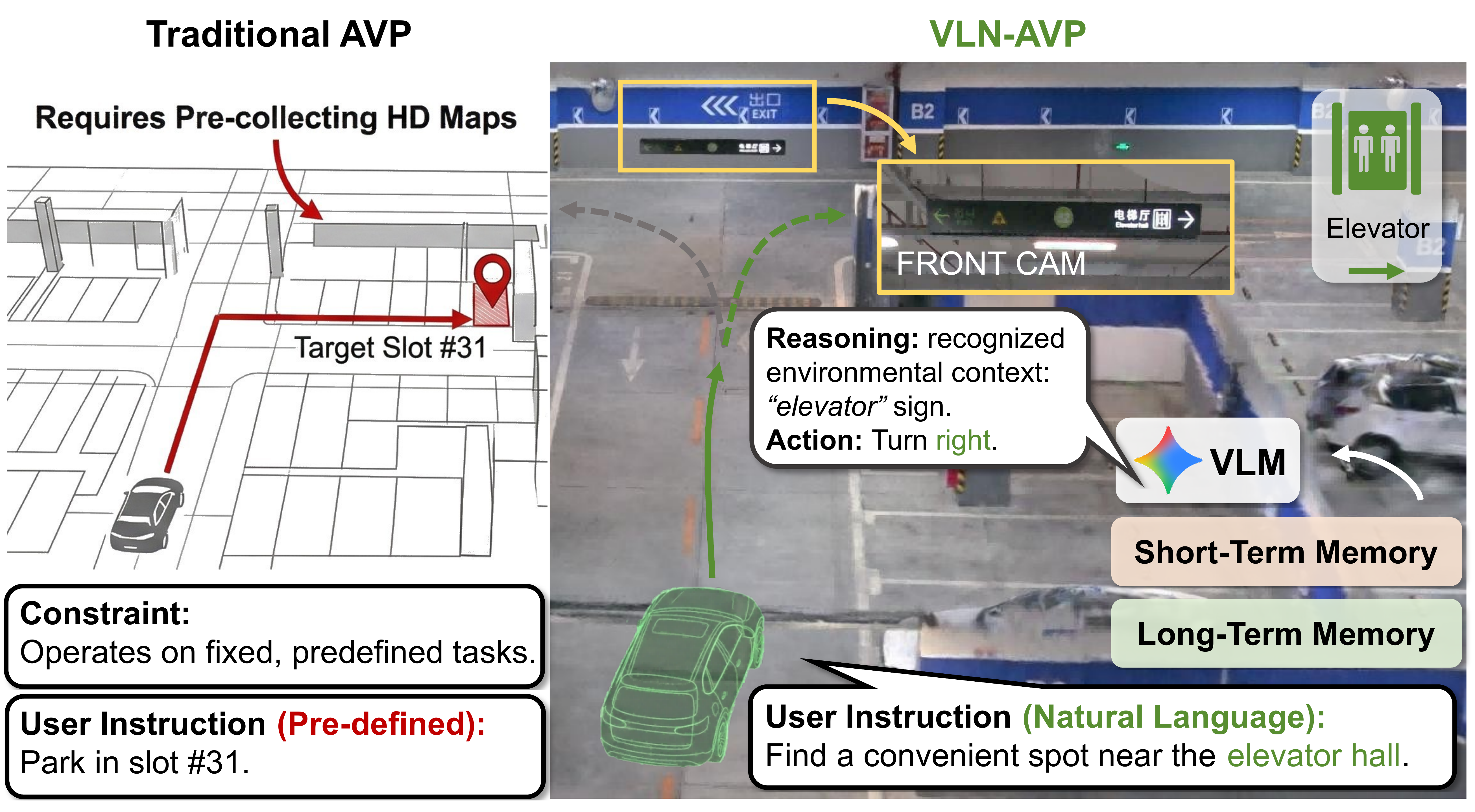}
        \caption{Traditional AVP systems require pre-built high-definition (HD) maps and are limited to pre-defined tasks. In contrast, VLN-AVP is capable of comprehending semantic cues in the surrounding environment and making driving decisions by a VLM-based navigation framework. Interpreting the human instructions, the system can find the way to the demanded parking space near the elevator. }
        \label{fig:introduction}
    \end{figure}

    They cannot handle more complex instructions like ``navigate to the exit" or ``park near the elevator lobby". To perform more flexible and intelligent AVP tasks, the AVP system must (1) comprehend natural language instructions, (2) interpret semantic environmental contexts, and (3) operate independently without pre-built maps.

    Recent studies in Autonomous Driving (AD) \cite{tian2024drivevlm, sima2024drivelm, ma2024lampilot, wen2024dilu} have explored employing Vision Language Models (VLMs) to enhance autonomous driving performance by fine-tuning or zero-shot methods. However, both approaches face significant challenges when applied to AVP tasks. This is because these methods are designed primarily for high-speed driving on open roads, making it difficult to generalize them directly to the complex, enclosed environments of underground parking garages. 

    Meanwhile, the successful application of zero-shot VLMs in Vision Language Navigation (VLN) \cite{gu2022vln} inspires us to apply VLMs to AVP tasks. VLN requires VLM-powered agents to comprehend natural language instructions and navigate to targets in unseen environments. The AVP scenario, with its low speed and enclosed environment, presents an ideal case for zero-shot VLM methodologies. Moreover, VLM can perceive critical environmental cues in parking lots, such as signs and ground markings, and subsequently make autonomous-cruising decisions based on these scene descriptions. Therefore, leveraging the strong capabilities of VLMs, AVP can be achieved in previously unseen parking scenarios, eliminating the reliance on prior environmental knowledge, such as pre-built maps.

    In this work, we propose VLN-AVP, a zero-shot vision-language navigation framework for AVP. Our approach combines the precise spatial perception of the Bird's-Eye-View (BEV) representation with the general reasoning capabilities of VLMs. To ensure stable and efficient navigation, we introduce a hybrid memory system: (1) a short-term perception memory that tracks semantic visual cues to augment the VLM's immediate decision-making, and (2) a long-term topological memory that facilitates stable policy learning by integrating past experiences. As shown in Figure \ref{fig:introduction}, the incorporation of VLMs allows the AVP system to interpret diverse natural language instructions, utilize semantic environmental context, and operate independently of pre-built maps. We also propose an underground parking dataset and benchmark to compensate for the lack of existing benchmarks. Finally, we conduct extensive experiments in both high-fidelity simulation and real-world parking scenarios, demonstrating that our framework significantly outperforms existing VLN and autonomous driving methods in both navigation success and efficiency.

    The contributions of this work include:
    \begin{itemize}
        \item We propose VLN-AVP, a zero-shot vision-language navigation framework for AVP. Compared to traditional methods, it requires no pre-built map, interprets semantic visual contexts in parking scenarios, and enables intuitive navigation using natural-language instructions. 

        \item We introduce a hybrid memory architecture. A short-term memory augments VLM's decision-making by retaining semantic visual cues, resolving the instability of VLM's single-frame decision-making in existing methods. A long-term topological memory facilitates stable policy learning through past experiences, improving efficiency and stability compared to other approaches. 

        \item We construct the VLN-AVP dataset and benchmark. Featuring 10 high-fidelity parking scenes and over 1,000 navigation episodes, it has the largest number of garage scenes to date and is the first VLN benchmark for underground parking. 

        \item Extensive experiments conducted on both simulators and real-world vehicles proved the outstanding performance of our method compared to existing approaches. 
    \end{itemize}

    \section{Related Works}

    \subsection{Methods for Autonomous Valet Parking}
    Early works on AVP \cite{avp-structure, avp-emobile} primarily focused on building a complete system to conduct real-world experiments, thereby proving the feasibility of the overall method. Recently, more and more research has focused on a specific part of the AVP system, especially on mapping and localization methods. Semantic SLAM-based approaches \cite{avp-slam, liu2024less, InversePerspective} used robust semantic features segmented from the inverse perspective mapping (IPM) image to build maps of underground parking slots. AVP-loc \cite{avp-loc} proposed a novel data association method to find the relation between current observations and the vectorized HD map, enabling stable localization performance in the underground parking scenarios.

    To sum up, current AVP algorithms heavily rely on pre-built maps, whereas our approach is designed to eliminate this dependency.

    \subsection{Vision Language Navigation}
    In vision-language navigation (VLN), an agent follows human instructions to navigate in previously unseen environments \cite{gu2022vln}. Nowadays, VLN methods primarily encompass two methodological paradigms: training-based approaches \cite{chen2022duet, zhang2024navid, cheng2024navila} and zero-shot approaches \cite{zhou2024navgpt, chen2024mapgpt, zhang2025apexnav}. Some methods \cite{zhang2024navid, cheng2024navila} leverage the capabilities of VLMs and fine-tune selected layers' parameters to adapt to navigation tasks. Benefiting from the enhanced performance of large language models (LLMs) and VLMs, zero-shot VLN approaches \cite{zhou2024navgpt, chen2024mapgpt} have also achieved excellent performance across diverse scenarios. Specifically, ReasonNav \cite{chandaka2025reasonnav} proposed integrating detection models with VLMs to identify environmental signage and other information. 

    However, as there are fundamental differences in operational scenarios, navigation rules, and control methodologies between mobile robots and AVP, VLN approaches face a large gap in AVP applications. We therefore propose a vision-language navigation framework specifically designed for AVP, enabling autonomous navigation from human instructions across diverse parking environments.

    \begin{figure*}[!t]
        \centering
        \includegraphics[width=0.85\textwidth]{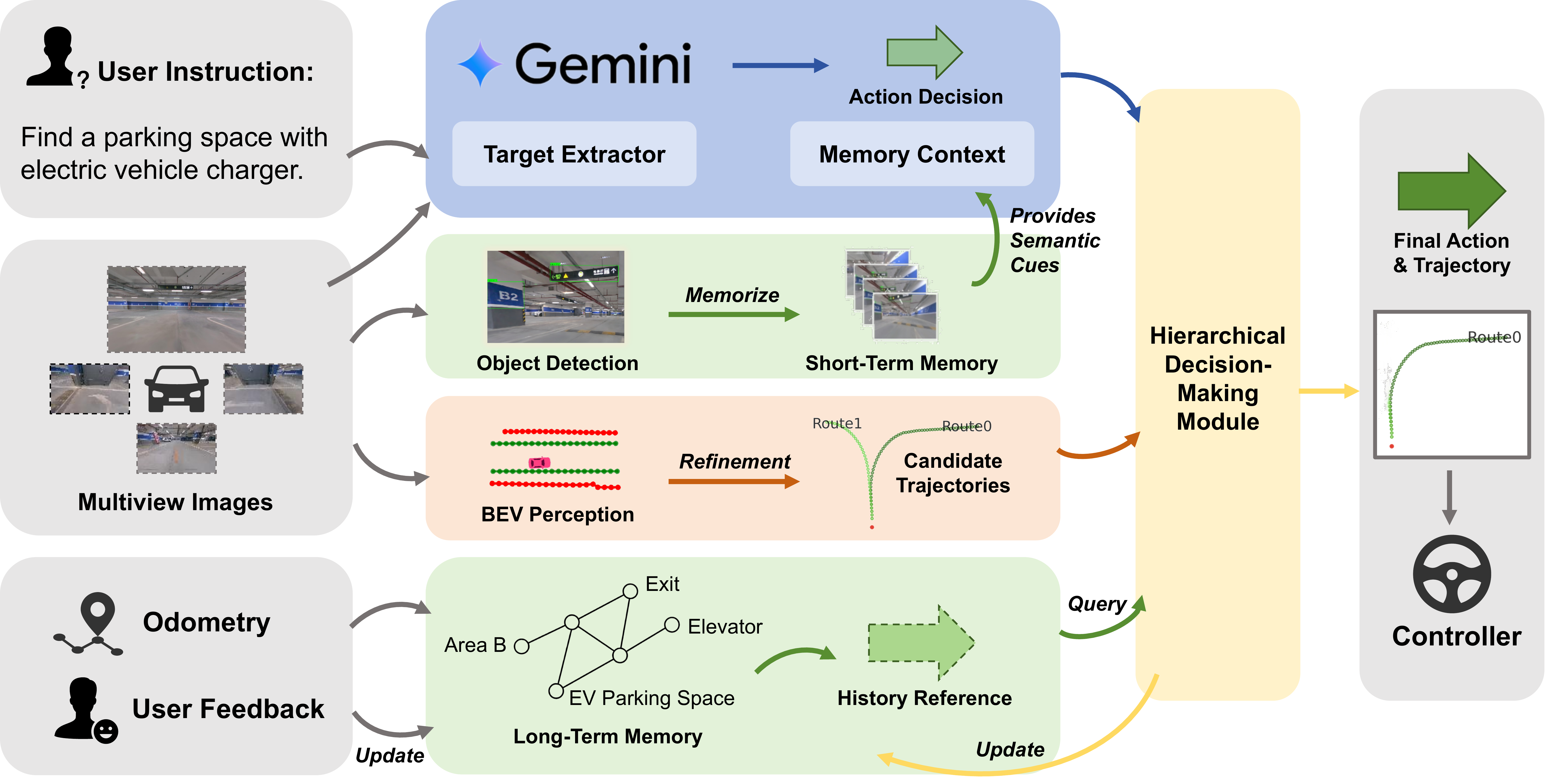}
        \caption{ The structure of VLN-AVP. The orange block updates at high frequency to provide low-level BEV perception and candidate trajectories. The blue block is the VLM core, which performs high-level, low-frequency decision-making based on current observation, short-term memory, and human instructions. The green blocks are the long-short-term memory modules. Finally, the hierarchical decision-making module bridges all other modules and chooses the most appropriate trajectory. 
        }
        \label{fig:framework}
    \end{figure*}

    \subsection{Vision Language Model for Autonomous Driving}
    With the enhanced performance of VLM, VLM-based autonomous driving methods have
    been widely studied to address corner cases that remained in traditional approaches. Recent works \cite{tian2024drivevlm, mao2023gpt, xu2024drivegpt4, fu2025orion} leveraged LLM or VLM as the backbone of an end-to-end network and performed fine-tuning on large-scale datasets. However, data-driven approaches often entail prohibitive training costs and still struggle to generalize across diverse, unstructured scenarios. Other works \cite{qiao2025lightemma, fu2024drive, ma2024lampilot, wen2024dilu} relied on VLMs to perform zero-shot decision-making during autonomous driving. These methods also face generalization challenges in underground parking scenes and usually cannot be applied directly to closed-loop tests or real vehicles. 

    In summary, while current VLM-based systems either suffer from high training costs or lack the robustness required for zero-shot deployment in parking environments, our VLN-AVP aims to bridge this gap. It provides a zero-shot framework that achieves reliable closed-loop navigation in unstructured parking scenarios.

    \newcolumntype{C}{>{\centering\arraybackslash}X}

    \begin{table}[!t]
        \centering
        \caption{Comparison of Existing Methods and VLN-AVP}
        \label{tab:related_works_compare}
        \setlength{\tabcolsep}{2pt}
        \scriptsize
        \begin{tabularx}{\columnwidth}{@{}CCCC@{}}
            \toprule 
            Method type and representative works & Fits underground parking scenarios & Generates drivable trajectories & Open-target navigation \\
            \midrule 
            AVP Methods \cite{avp-structure, avp-slam}& $\checkmark$ & $\checkmark$ & $\times$ \\
            \midrule
            VLN Methods \cite{chen2024mapgpt, chandaka2025reasonnav}& $\times$ & $\times$ & $\checkmark$ \\
            \midrule
            VLM-based AD Methods \cite{ma2024lampilot, wen2024dilu}& $\times$ & $\checkmark$ & $\checkmark$ \\
            \midrule
            \textbf{VLN-AVP (Ours)} & $\checkmark$ & $\checkmark$ & $\checkmark$ \\
            \bottomrule
        \end{tabularx}
    \end{table}

    \section{Methodology}

    We propose VLN-AVP, a zero-shot vision-language navigation approach for autonomous valet parking. As depicted in Figure \ref{fig:framework}, the system comprises four core components: (1) a high-frequency, low-level module for spatial awareness and candidate trajectory generation, (2) a high-level, low-frequency decision core powered by VLM, (3) a hybrid long-short-term memory module, comprising a short-term memory which augments VLM's decision-making by retaining semantic visual cues, and a long-term topological memory which facilitates stable policy learning through past experiences, and (4) a hierarchical decision-making module that bridges the high and low frequency modules, and makes the final driving decisions. 

    \subsection{BEV Perception and Candidate Trajectory Generation}

    The foundation of our system is its ability to perceive the local
    environment without relying on pre-built HD maps. We employ MapTR \cite{liao2022maptr}, a transformer-based BEV perception model, to obtain low-level road structures at a high frequency. It processes synchronized images from four surround-view cameras mounted on the vehicle $\mathcal{I}_{t} = \{I_{t}^{\text{front}}
    , I_{t}^{\text{rear}}, I_{t}^{\text{left}}, I_{t}^{\text{right}}\}$ and then directly outputs vectorized representations of key map elements in the BEV space. For the AVP task, we primarily focus on its prediction of lane centerlines, $\mathcal{L}_{t} = \text{MapTR}(\mathcal{I}
    _{t})$.

    These raw predictions of centerlines are then passed to a trajectory optimizer, which combines them with RS curves \cite{reeds1990optimal} and the vehicle's kinematics model, and generates a set of drivable trajectories $\mathcal{T}$:
    \begin{equation}
        \mathcal{T}_{t} = f_{\text{optim}}(\mathcal{L}_{t}) = \{\tau_{i}\}_{i=1}^{N_t},
    \end{equation}
    where $N_t$ is the number of candidate trajectories at time $t$. Each $\tau_{i}$ is formally defined as an ordered sequence composed of $M$
    two-dimensional waypoints in the BEV coordinates:
    \begin{equation}
        \tau_{i} = \{ p_{j} \}_{j=1}^{M}, \text{ where }p_{j} \in \mathbb{R}^{2}.
    \end{equation}
    Each trajectory $\tau_{i}$ represents a potential path for the vehicle, such
    as continuing straight, turning left, or turning right at a junction. This
    set $\mathcal{T}$ serves as the action space for the downstream controller of the vehicle.

    \subsection{Detection and Short-Term Memory}
    Many methods \cite{ma2024lampilot, chen2024mapgpt, chandaka2025reasonnav} have employed VLMs to perform high-level driving or navigation decisions, but they typically rely on single-frame image or text as input, which has several limitations: (1) VLM usually runs at a low frequency, so it's very possible to miss some critical visual information, such as a sign that appears only for a short time. (2) VLMs cannot have a space-time perception of the environment with a single image. 
    
    To address these limitations, we employ Grounding DINO \cite{liu2024grounding}, a high-frequency object detection module, to capture visual navigation cues from $I^{\text{front}}_t$. Then, we design a short-term memory (STM) module to retain these perceptions and compensate for the low-frequency of VLM. In the AVP scenario, we use ``sign" as the prompt for detection, as signs are the main source of navigation information. Each detection result $B^{\text{detect}} = \text{DINO}(I^{\text{front}}_t)$ is represented as an image with bounding boxes, and together with its timestamp $t$, bounding boxes $B$ and the vehicle's yaw angle $\theta$, we form a frame of the STM $m = (I^\text{detect}, B, t, \theta)$. The STM is a $K_{\text{STM}}$-frame queue, denoted as $\text{STM} = \{m_j\}_{j=1}^{K_{\text{STM}}}$. When a new frame $m_\text{new}$ is added to the STM, 3 filtering steps are performed to remove frames that are no longer necessary: 
    \begin{itemize}
        \item \textbf{Object Filter: }Try to match $m_\text{new}$ with other frames in STM by IoU-based object tracking. Replace old frames with the same object(s) in $m_\text{new}$. This step eliminates redundant frames of the same objects caused by the high-frequency detection. 
        \item \textbf{Directional Filter: }Remove frames with significantly different yaw angles from the current vehicle pose, i.e., $|\theta_{new} - \theta_{old}| > \Delta \theta_{max}$. This step ensures that the sign directions in the STM are consistent with the current vehicle pose. 
        \item \textbf{Temporal Filter: }Remove frames that are too old, i.e., $t_{new} - t_{old} > \Delta t_{max}$. If the queue is still full, the oldest frame is removed. 
    \end{itemize}

    \subsection{High-level Decision-making by Vision-Language Model}

    To enhance our system with semantic understanding and zero-shot generalization
    capabilities, we leverage the power of large-scale VLMs. When a user issues a
    natural language command, such as ``park near the elevator entrance of level B2''.
    A target extractor first decomposes this instruction into a sequence of
    actionable navigation sub-goals:
    \begin{equation}
        G = f_{\text{extract}}(L_{\text{user}}) = (g_{1}, g_{2}, \dots, g_{K_g}).
    \end{equation}
    Here, $G$ is a sequence of length $K_g$, where each sub-goal $g_{i}$ is a noun describing a sub-goal in the overall instruction (e.g., ``level B2'' or ``elevator entrance'').

    The core of the guidance system is the VLM module. At each decision step, the VLM receives a compound prompt $\mathcal{P}$:
    \begin{equation}
        \mathcal{P}= [I_{front} \oplus \{I_{detect, j}\}_{j=1}^{K_{\text{STM}}} \oplus S \oplus g],
        \label{equ:prompt}
    \end{equation}
    where $I_{front}$ is the current forward-facing camera image, $\{I_{detect, j}\}_{j=1}^{K_{\text{STM}}}$ is the collection of detection results in the STM, $S$ is a system prompt defining the navigation task, and $g$ is the current navigation sub-goal.
    The VLM then infers the appropriate high-level meta-action:
    \begin{equation}
        a_\text{vlm}= \text{VLM}(\mathcal{P}),
    \end{equation}
    where $a_\text{vlm}\in \mathcal{A}= \{ \text{forward, left, right, stop}\}$ is extracted from the VLM's output. If $a_\text{vlm}$ is the stop action, the navigation ends. Otherwise, $a_\text{vlm}$ is cached in an action cache $A_{cache}$ in the hierarchical decision module, and later sent to the low-level controller.

    \begin{figure}[!t]
    \centering
    \includegraphics[width=.95\linewidth]{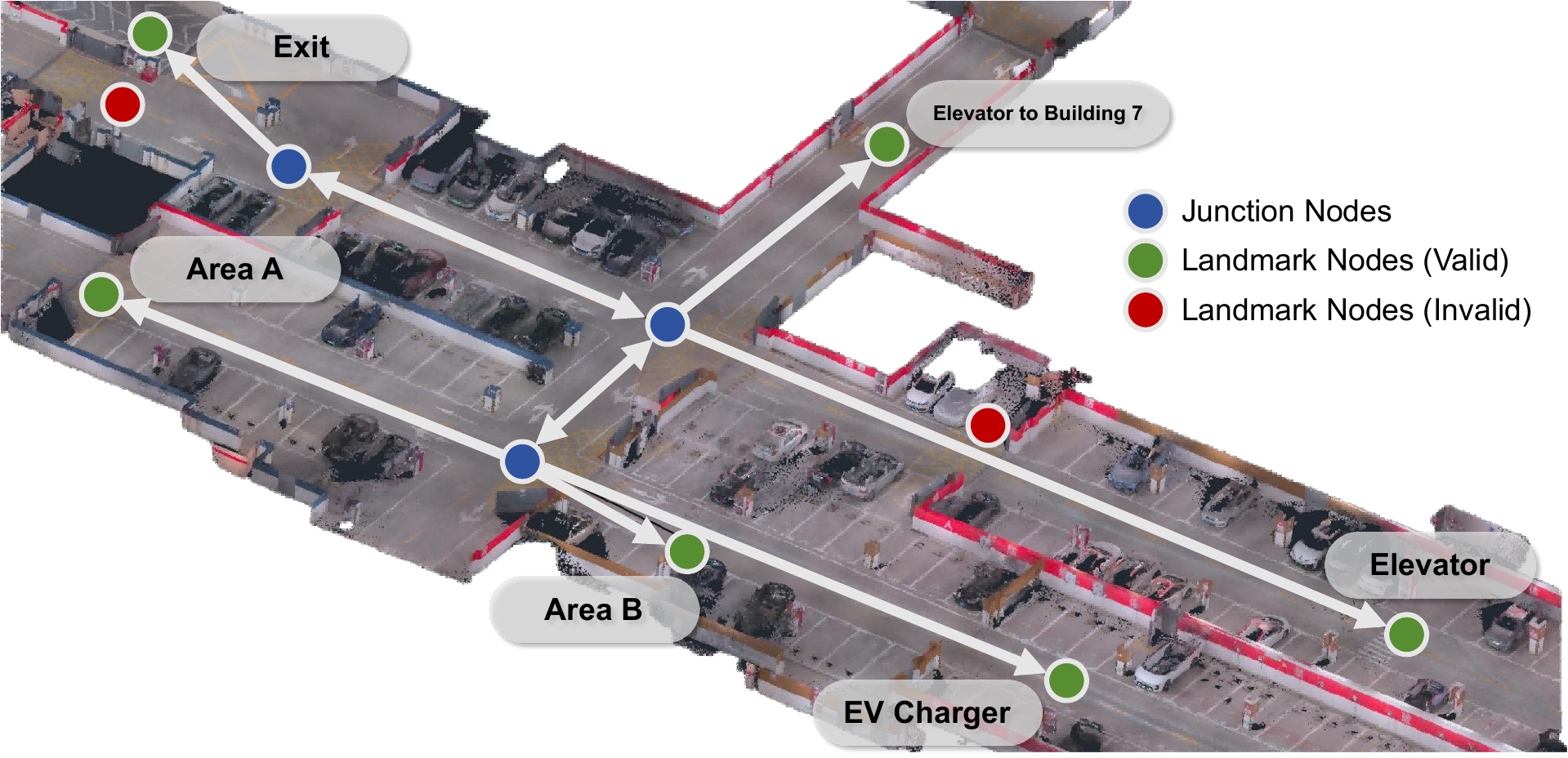}
    \caption{Visualized LTM on a simulated scene after 10 navigation episodes. }
    \label{fig:ltm_show}
    \end {figure}

    \subsection{Long-Term Topological Memory}
    VLMs and short-term memory work well in zero-shot scenarios, such as navigating a public garage for the first time. But for repeated tasks in the same garage (like daily commutes), it is more efficient to leverage the historical experience of successful navigations. Therefore, we introduce a long-term topological memory (LTM). This topological graph gradually builds a sparse, stable representation of the environment from past experiences, compensating for the lack of a prebuilt map. Figure \ref{fig:ltm_show} shows an example of the LTM. 

    \textbf{Structure: }The LTM is represented as a directed graph $\mathcal{G}= (\mathcal{V}, \mathcal{E})$, consisting of: 
    \begin{itemize}
        \item \textbf{Junction Nodes:} $n_{\text{junc}}\in \mathcal{V}$ are defined as $n_{\text{junc}}= (p_{\text{junc}}, v)$, where $p_{\text{junc}} \in \mathbb{R}^{2}$ is the coordinate of the road junction center and $v \in \{0, 1\}$ is a validity bit. When the system is at an unseen junction, it initializes a junction node and calculates $p_{\text{junc}}$ from BEV perception. 

        \item \textbf{Landmark Nodes:} $n_{\text{lm}}\in \mathcal{V}$ are defined as $n_{\text{lm}}= (p_{\text{lm}}, v)$. It is recorded at task completion points when the system executes a `stop' action. 

        \item \textbf{Edges:} $e \in \mathcal{E}$ are defined as $e=(n_{\text{source}}, n_{\text{destination}}, \mathbf{d}, v)$. It is established between nodes when the vehicle drives from one node $n_{\text{source}}$ to another $n_{\text{destination}}$. The system will record the trajectory $\{ p_{j} \}_{j=1}^{M}$ of length $M$ chosen at $n_{\text{source}}$. Then the direction vector of the edge $\mathbf{d} \in \mathbb{R}^{2}$ is calculated by the following equation:
    \begin{equation}
        \mathbf{d} = \frac{p_{M}- p_{M-4}}{\| p_{M}- p_{M-4}\|_{2}}.
        \label{equ:direction}
    \end{equation}
        This vector uses the 5th-to-last point to the final point to represent the direction of the edge. Storing a direction vector instead of a relative command (`left' / `right') makes the memory robust to the vehicle's different approach angles. 
    \end{itemize}

    \begin{figure}[!t]
    \centering
    \includegraphics[width=.95\linewidth]{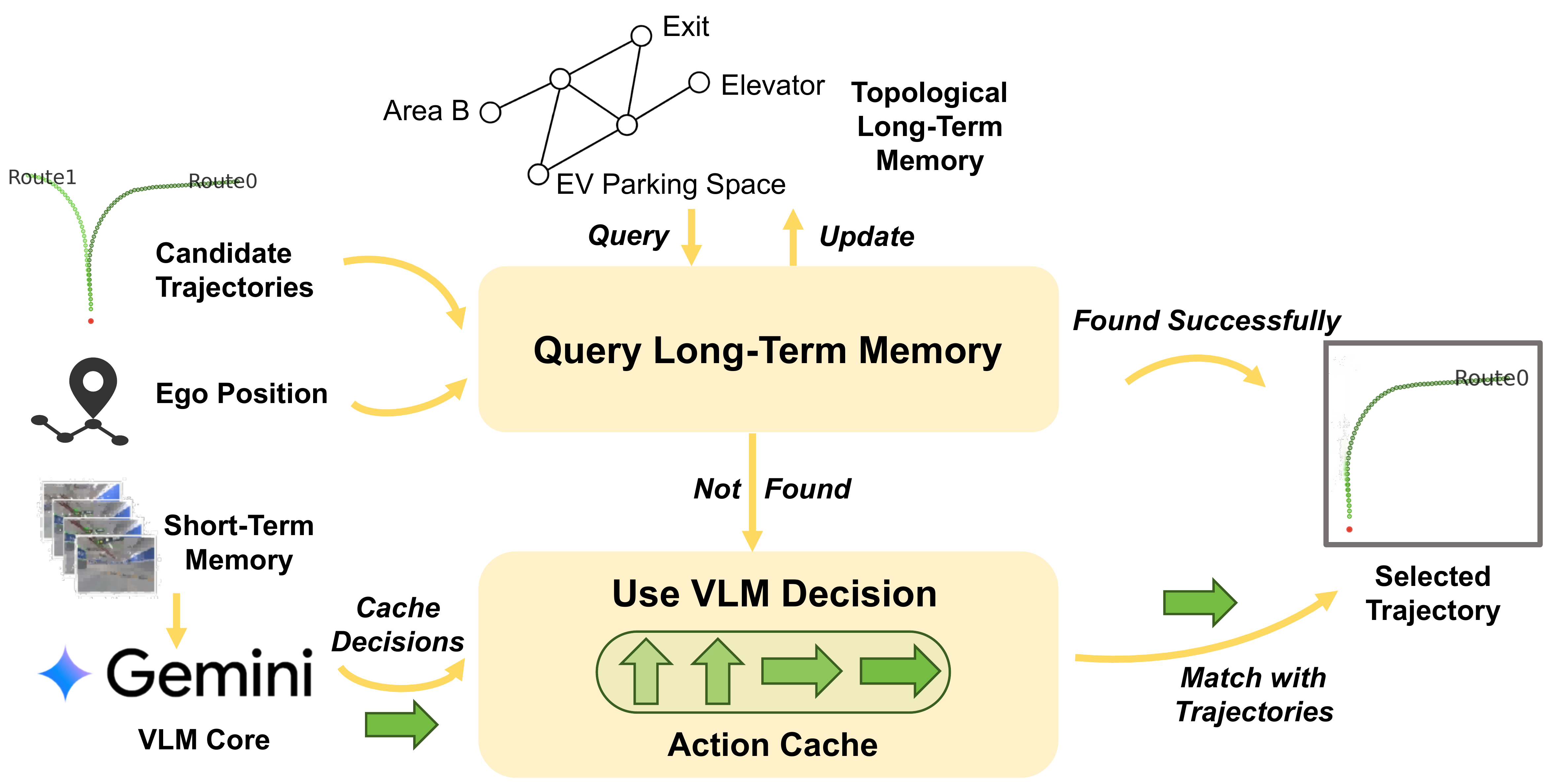}
    \caption{The framework of our proposed hierarchical decision-making module. Based on the long-term memory module and the VLM output, the candidate trajectory is selected in a hierarchical way.}
    \label{fig:query}
    \end {figure}

    \begin{figure*}[!t]
        \centering
        \includegraphics[width=.95\textwidth]{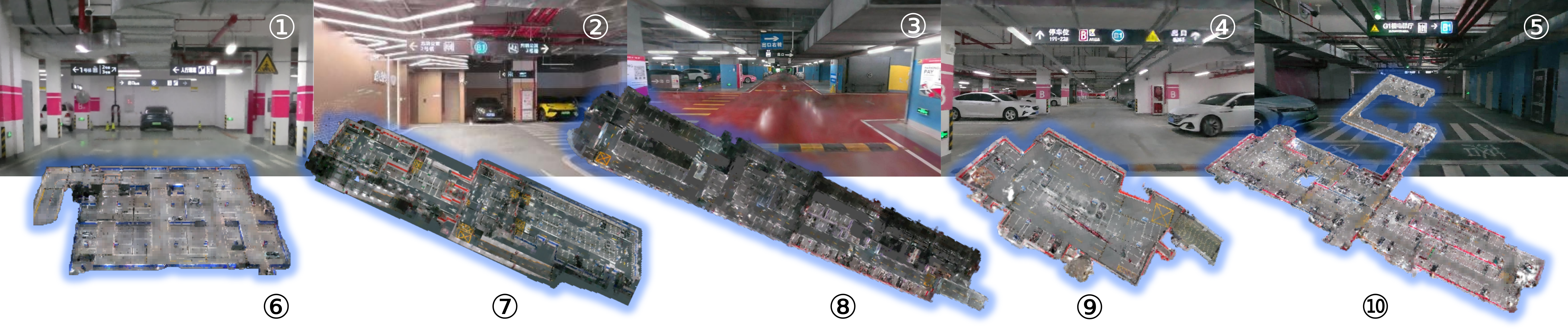}
        \caption{10 scenes in the VLN-AVP dataset. Top row: 3DGS renderings of 5 scenes; Bottom row: Top-down views of the remaining 5 scenes.}
        \label{fig:simulation_maps}
    \end{figure*}
    
    \textbf{Memory Updating:} After each navigation episode, node and edge validity bits $v \in \{0, 1\}$ are updated based on user feedback:
    \begin{itemize}
        \item \textbf{Navigation Success:} All traversed nodes and edges are set to $v=1$.

        \item \textbf{Navigation Failure/Continuation:} If the user indicates the navigation is unsuccessful when the vehicle stops, the landmark node is set to $v=0$. Upon the user's request, the system may continue the navigation. 

        \item \textbf{System Triggered Failure:} If the system follows the LTM memory but cannot find a valid trajectory, it implies there has been environment changes. In this case, no landmark node will be generated, and the validity bits of the edges leading to the failure will be set to 0. If in later episodes the edge recovers and is traversed, its validity bit will be set to 1. 

    \end{itemize}

    \subsection{Hierarchical Decision-Making Module}
    The hierarchical decision-making process (Figure \ref{fig:query}) intelligently balances long-term experience and short-term VLM guidance. When there is only one candidate trajectory, the system directly executes it and keeps updating $\mathcal{T}$. When the vehicle arrives at a decision point, such as a junction (i.e., $|\mathcal{T}| \ge 2$):

    \begin{enumerate}
        \item \textbf{Query Long-Term Memory:} If a valid landmark node that matches the current target exists, the system uses graph search to find a path $\{e_1, e_2, \dots, e_n\}$ to the target node. If successful, a direction vector $\mathbf{p}_{i}$ is calculated for each current candidate trajectory by Eq. (\ref{equ:direction}). The system then identifies the candidate trajectory $\tau_{i^*}$ with $\mathbf{p}_{i^*}$ that maximizes its cosine similarity with the direction vector of $e_1$ and satisfies the similarity threshold $\theta_{\text{threshold}}$. If $i^*$ is found, the system selects the trajectory $\tau_{i^*}$ for execution.

        \item \textbf{Query Cached VLM Decision:} During navigation, the decision module keeps querying the VLM according to Eq. (\ref{equ:prompt}) and updating the action cache $A_{cache}$. If no valid long-term memory is available for guidance, the system relies on the VLM's judgment by selecting the latest action in $A_{cache}$. The action is mapped to the most suitable $\tau_{i^*}$ based on angular deviation. This cache design decouples low-frequency VLM inference from the trajectory selection process, reducing decision-making latency.

        \item \textbf{Execution:} The selected trajectory $\tau_{i^*}$ is passed to the low-level vehicle controller for smooth execution. After the execution of $\tau_{i^*}$, the system updates $\mathcal{T}$ and repeats the decision-making process.
    \end{enumerate}

    In summary, the hierarchical strategy ensures efficiency in familiar areas while leveraging the VLM’s zero-shot generalization for novel situations.

    \section{Experiments}
    \subsection{VLN-AVP Dataset and Benchmark}
    A significant challenge in AVP research is the lack of high-quality simulation
    datasets for underground parking scenarios.
    To address this, we constructed the VLN-AVP dataset and benchmark.
    \subsubsection{Dataset Construction}
    We first used a handheld scanning device equipped with LiDAR and fisheye cameras
    to capture dense point cloud and image data from real-world underground parking
    garages. These datasets were then processed using 3D Gaussian Splatting (3DGS)
    \cite{kerbl20233dgs} techniques to generate photorealistic, large-scale, and
    continuous 3D scenes (Figure \ref{fig:simulation_maps}). 

    \begin{figure*}[!t]
        \centering
        \includegraphics[width=0.9\textwidth]{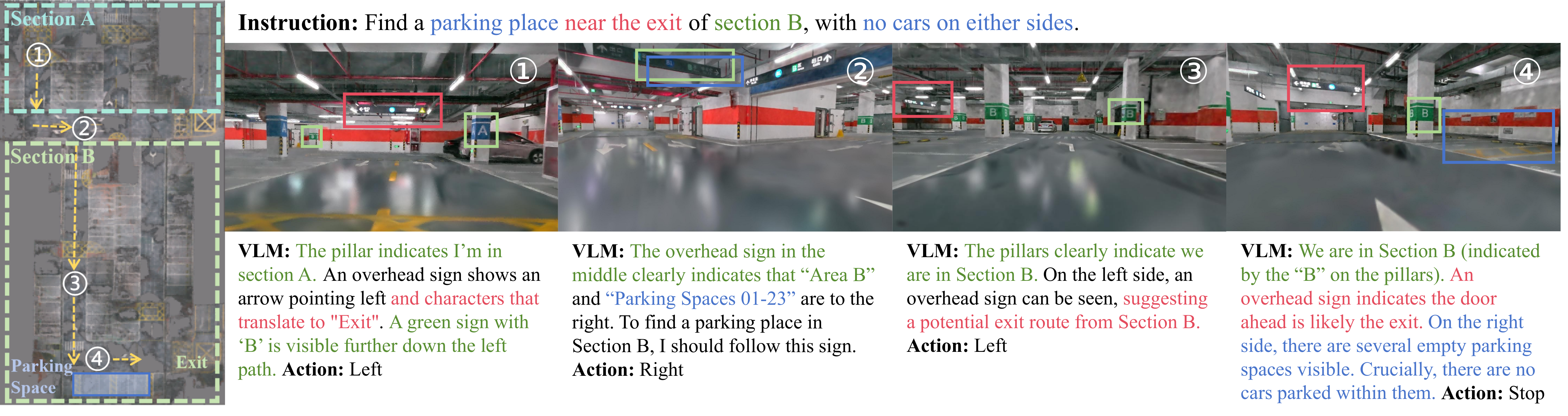}
        \caption{A successful case study of our VLN-AVP system navigating a complex,
        multi-stage instruction in simulation. The left-most panel shows the global trajectory. The right panels show the front camera view at key decision points, along with the VLM's reasoning. }
        \label{fig:sim_demo}
    \end{figure*}

    \begin{table}[!t]
        \centering
        \caption{Comparison of Different Underground Parking Datasets}
        \label{tab:dataset_compare}
        \resizebox{\columnwidth}{!}{%
        \begin{tabular}{lccc}
            \toprule Dataset & Number of Garages & Render Method & VLN Episodes \\
            \midrule SUPS \cite{hou2022sups}     & 1                 & Unity         & $\times$     \\
            SUSTech-COE \cite{wang2024occupancy} & 1                 & Unreal        & $\times$     \\
            GarageWorld \cite{cui2024letsgo}     & 8                 & 3DGS          & $\times$     \\
            GarageNet \cite{yue2025point}        & 2                 & Point Cloud   & $\times$     \\
            \textbf{VLN-AVP (Ours)}              & \textbf{10}       & \textbf{3DGS} & $\checkmark$ \\
            \bottomrule
        \end{tabular}
        }
    \end{table}

    As shown in Table \ref{tab:dataset_compare}, our dataset provides more diverse environments (10 garages) compared to previous works. More importantly, it is the first to provide specific VLN episodes in underground parking scenarios.

    \subsubsection{Benchmark Construction}

    We also developed VLN-AVP Benchmark, a complete benchmark for evaluating VLN tasks in our dataset. It has over 1000 navigation episodes across all 10 scenes, each including natural-language navigation instructions, starting location, target location, and the ground-truth trajectory. We created 2 subsets according to task complexity: 
    \begin{itemize}
        \item \textbf{VLN-AVP-Regular}: Target is a single, common location, like ``go to the exit" or ``go to the elevator entrance".
        \item \textbf{VLN-AVP-Challenging}: Target may contain combinations of additional demands, like ``find a parking space with electric vehicle charging port" or ``go to the elevator nearest to building B".
    \end{itemize}

    \subsection{Experiment Setup}

    \subsubsection{Implementation Details}
    Our system utilizes the Gemini-3-flash model as the VLM for decision-making. The BEV perception module is built upon MapTR \cite{liao2022maptr} to predict road structure. For our short-term memory, the cache size is set to $K_{\text{STM}}=6$. In simulated experiments, we use a customized simulator built upon 3DGS \cite{kerbl20233dgs}. Its design is based on the Real2Sim2Real simulator in REAP \cite{li2026reap}, with extensions for VLN task interfaces and realistic vehicle kinematics. Our method runs along with the simulation environment on a desktop with a single NVIDIA 4090 GPU. In real-world experiments, we deploy the system on a real vehicle equipped with an NVIDIA 2060 GPU. All experiments are conducted in continuous, real-time environments. 

    \subsubsection{Evaluation Metrics}
    We adopt standard metrics from VLN to quantitatively evaluate performance, including Success Rate (SR), Oracle Success Rate (OSR), Success weighted by Path Length (SPL) and Navigation Error (NE). An episode is considered successful if the agent stops within the bounding box of the target location or area. Simulated experiments are conducted on VLN-AVP-Regular and VLN-AVP-Challenging subsets. Real-world experiments are conducted on 30 navigation trials in 2 real underground parking garages.

    \subsubsection{Baseline}
    To the best of our knowledge, no prior work has specifically addressed VLN in
    underground parking environments. So we compare our method against 2 relevant types of methods:
    \begin{itemize}
        \item \textbf{VLN Methods}: Including two representative works in zero-shot VLN, MapGPT \cite{chen2024mapgpt} and ReasonNav \cite{chandaka2025reasonnav}. For MapGPT, we created an enhanced version, MapGPT w/ BEV, where waypoints are sampled along the reference trajectories provided by the BEV perception module. ReasonNav is natively designed for all-directional movement, so we use the original method.

        \item \textbf{Autonomous Driving (AD) Methods}: Including two representative VLM-based works in autonomous driving, LaMPilot \cite{ma2024lampilot} and DiLu \cite{wen2024dilu}. The same BEV perception and other inputs are provided to these two methods. 
    \end{itemize}
    Besides, all methods use the same VLM core as VLN-AVP. By providing baselines with the same BEV understanding and other inputs as our method, we enable a fair comparison of other components, such as decision-making and the memory module. 


    \begin{table}[!t]
        \centering
        \caption{Simulation Experiment Results}
        \setlength{\tabcolsep}{1mm}
        \begin{tabular}{@{}lcccc@{}}
            \toprule Method & SR (\%) $\uparrow$ & OSR (\%) $\uparrow$ & SPL (\%) $\uparrow$ & NE (m) $\downarrow$ \\
            \midrule \multicolumn{5}{c}{\textit{Regular Tasks}}  \\
            \midrule
            \multicolumn{5}{l}{\textbf{VLN Methods}} \\
            MapGPT \cite{chen2024mapgpt} (ACL'24) & 9.3& 18.3& 3.5& 39.8\\
            MapGPT w/ BEV & 37.6& 46.0& 26.6& 26.7\\
            ReasonNav \cite{chandaka2025reasonnav} (CoRL'25) & 23.0& 28.2& 13.6& 28.2\\
            \midrule
            \multicolumn{5}{l}{\textbf{AD Methods}} \\
            LaMPilot \cite{ma2024lampilot} (CVPR'24) & 49.7& 54.4& 34.8& 19.5\\
            DiLu \cite{wen2024dilu} (ICLR'24) & 47.4& 50.2& 31.5& 22.6\\
            \textbf{VLN-AVP (Ours)}& \textbf{65.2}& \textbf{74.6}& \textbf{43.1}& \textbf{13.2}\\
            \midrule \multicolumn{5}{c}{\textit{Challenging Tasks}} \\
            \midrule
            \multicolumn{5}{l}{\textbf{VLN Methods}} \\
            MapGPT w/ BEV & 19.3& 26.0& 14.8& 31.8\\
            ReasonNav \cite{chandaka2025reasonnav} & 17.6& 22.7& 10.7& 34.5\\
            \midrule
            \multicolumn{5}{l}{\textbf{AD Methods}} \\
            LaMPilot \cite{ma2024lampilot} & 33.6& 40.3& 24.7& 27.8\\
            DiLu \cite{wen2024dilu} & 30.2& 41.1& 23.5& 29.2\\
            \textbf{VLN-AVP (Ours)}& \textbf{45.4}& \textbf{59.7}& \textbf{33.2}& \textbf{18.4}\\
            \bottomrule
        \end{tabular}

        \label{tab:main_results}
    \end{table}

    \subsection{Results in Simulated Environments }
    The quantitative results are summarized in Table~\ref{tab:main_results}. Our
    VLN-AVP demonstrates a substantial improvement over all other baselines. We analyzed some key reasons for the performance gap. One primary reason is that our hybrid memory design is more reliable and efficient for AVP scenarios. In contrast, image-based memory in ReasonNav and text-based graph memory in MapGPT are less robust to the complex environments of parking lots in the long term, while few-shot learning methods in DiLu and LaMPilot are inefficient at providing VLM with the necessary information in the short term. In addition, VLN methods often make frequent U-turns, which are kinematically infeasible for a real-world vehicle. Our VLN-AVP, constrained by the kinematically-aware reference trajectories, naturally avoids this problem.

    To provide a more intuitive understanding of our system, we present a case study of a complex navigation task in Figure~\ref{fig:sim_demo}. This example demonstrates our method's generalization capabilities, which are unattainable by conventional AVP approaches.

    \begin{figure*}[!t]
        \centering
        \includegraphics[width=1\textwidth]{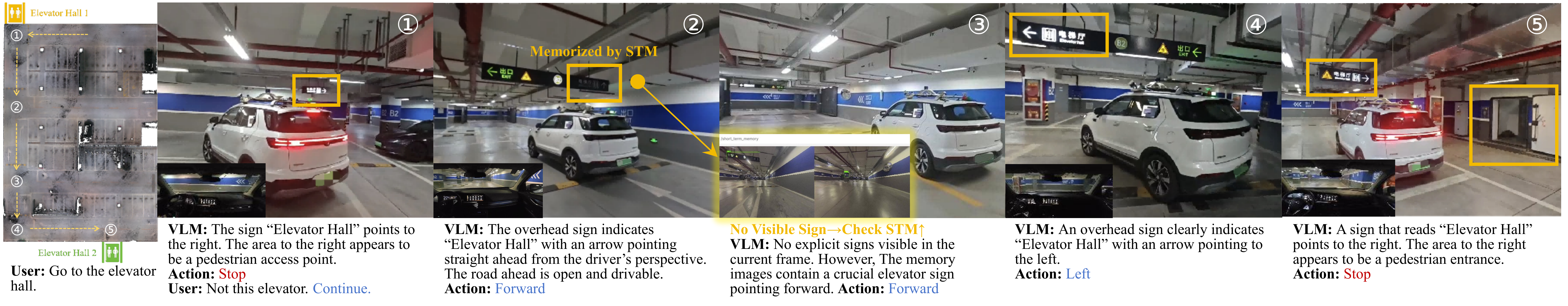}
        \caption{A qualitative experiment on a real-world vehicle platform. The system
        initially navigates to an incorrect elevator hall but successfully corrects its
        path and reaches the final destination after receiving user feedback. The middle picture shows how the short-term memory helps the VLM when there's no visual cue in the current frame. }
        \label{fig:real_world_test}
    \end{figure*}

    \subsection{Results in Real-world Environments}

    \begin{table}[!t]
      \centering
      \caption{Real-world Experiment Results}
      \begin{tabular}{lcc}
        \toprule
        Method & SR (\%) $\uparrow$ & OSR (\%) $\uparrow$ \\ \midrule
        LaMPilot \cite{ma2024lampilot} &  36.7&  40.0\\ 
        Ours (w/o LTM) &  53.3&  70.0\\
        \textbf{Ours} & \textbf{63.3}& \textbf{73.3}\\ 
        \bottomrule
      \end{tabular}
      \begin{minipage}{0.9\linewidth}
            \small
            \vspace{1ex}
            `w/o LTM' is without Long-Term Memory. 
       \end{minipage}
    \end{table}

    We also deployed our VLN-AVP on a real-world vehicle platform. As VLN methods are not kinematically feasible on real vehicles, we compare with LaMPilot \cite{ma2024lampilot}. Although real-world experiments contain fewer episodes, which diminishes the advantage of long-term memory, our method still outperforms the baseline, demonstrating robust zero-shot capability. 

    Figure~\ref{fig:real_world_test} illustrates a representative case study. Two
    elevators exist in the scene, while only one is the intended target of the
    user. VLN-AVP stops at the unwanted one first and continues to navigate to
    the correct one on request by the user's feedback. This capability for human-machine interaction enables error correction based on human feedback, which is also essential for real-world applications. The middle picture also shows how the short-term memory helps the VLM when there's no visual cue in the current frame. 

    \subsection{Ablation Studies}

    To validate the performance of our proposed memory components, we conducted a series of ablation studies.

    \begin{table}[!t]
        \centering
        \caption{Ablation Study of the Memory Components}
        \setlength{\tabcolsep}{1mm}
        \begin{tabular}{@{}lcccc@{}}
            \toprule Method & SR (\%) $\uparrow$ & OSR (\%) $\uparrow$ & SPL (\%) $\uparrow$ & NE (m) $\downarrow$ \\
            \midrule \multicolumn{5}{c}{\textit{Regular Tasks}}  \\
            \midrule
            VLN-AVP              & 65.2& 74.6& 43.1& 13.2\\
            w/o LTM              & 56.8& 70.4& 36.5& 15.6\\
            w/o LTM \& STM       & 42.2& 49.3& 30.9& 23.2\\
            \midrule \multicolumn{5}{c}{\textit{Challenging Tasks}} \\
            \midrule
            VLN-AVP              & 45.4& 59.7& 33.2& 18.4\\
            w/o LTM              & 38.7& 52.9& 28.5& 24.1\\
            w/o LTM \& STM       & 31.1& 39.5& 22.6& 29.9\\
            \bottomrule
        \end{tabular}
        \begin{minipage}{0.9\linewidth}
            \small
            \vspace{1ex}
            `w/o LTM' is without Long-Term Memory; `w/o LTM \& STM' is without Long- or
            Short-Term Memory.
        \end{minipage}
        \label{tab:ablation_study}
    \end{table}

    \subsubsection{Impact of Long-Short-Term Memory}
    We evaluated two variants of our system: one without the long-term memory (w/o
    LTM), and one without either the long-term or short-term memory (w/o LTM \&
    STM). The results are presented in Table~\ref{tab:ablation_study}.

    The data shows that each memory component contributes positively to the
    overall performance. 

    \subsubsection{Learning Ability of Long-Term Memory}
\begin{figure}[!t]
        \centering
        \includegraphics[width=\columnwidth]{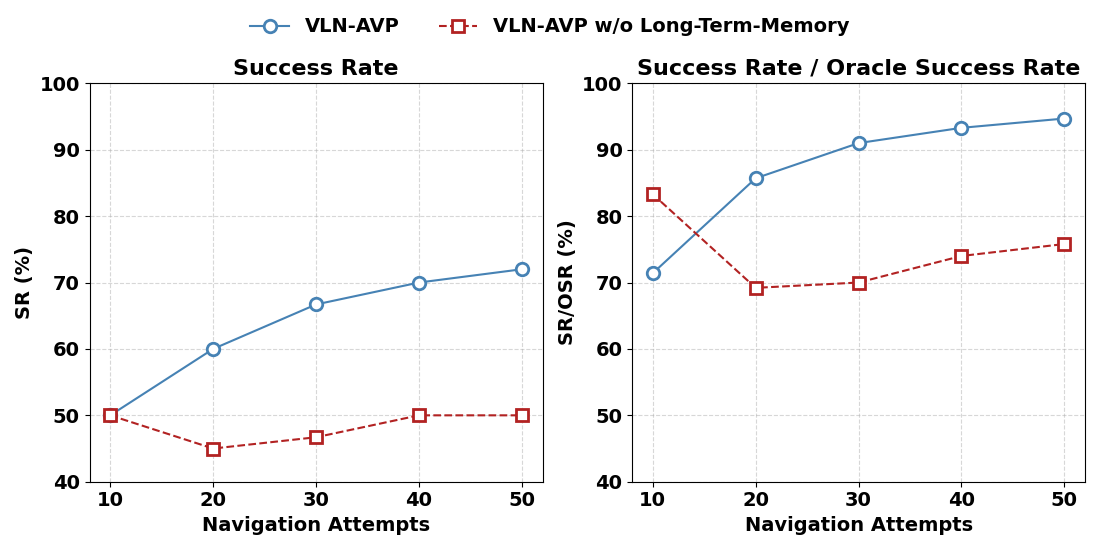}
        \caption{Performance comparison over 50 navigation attempts on one scene. }
        \label{fig:long_term_learning}
    \end{figure}

    To further investigate the long-term memory, we evaluated the
    system's performance over 50 navigation attempts in one scene. As
    illustrated in Figure~\ref{fig:long_term_learning}, the full VLN-AVP system
    exhibits a steady increase in both SR and the SR/OSR ratio. The rising SR demonstrates the system's ability to learn from past successful trials. More importantly, the increasing SR/OSR ratio indicates that the agent gradually learns to stop at the precise location, thanks to the updates of landmark nodes in the topological map. In contrast, the variant without long-term memory shows no clear learning trend.

    \subsubsection{Long-Term Memory and Cost}

    \begin{table}[!t]
      \centering
      \caption{Long-Term Memory and Cost}
	  \label{tab:ltm_cost}
      \begin{tabular}{lccc}
        \toprule
        Method & VLM Calls & Avg. Time (sec) & Cost (\$) \\ \midrule
        VLN-AVP (w/o LTM) & 1403 & 262.7 & 1.06 \\
        VLN-AVP & 317 & 161.7 & 0.24 \\ 
        \bottomrule
      \end{tabular}
      \begin{minipage}{0.9\linewidth}
            \small
            \vspace{1ex}
            `w/o LTM' is without Long-Term Memory. 
			`VLM Calls' is the total number of VLM calls made in 10 episodes; `Avg. Time' is the average navigation time per episode; `Cost' is total cost of API service in USD.
       \end{minipage}
    \end{table}

    To further understand the effect of long-term memory, we compare our full method with the one without long-term memory in terms of the total number of VLM calls, average navigation time per episode, and total API service cost over 10 episodes. The results in Table \ref{tab:ltm_cost} demonstrate that the long-term memory significantly reduces both the computational overhead and the financial cost. By retrieving navigation knowledge from the topological map, the system avoids redundant VLM reasoning in previously visited areas, resulting in fewer API calls and improved navigation efficiency.

    \subsubsection{Short-Term Memory and Decision-making}

    \begin{table}[!t]
      \centering
      \caption{Short-Term Memory and Decision-making}
      \begin{tabular}{lcc}
        \toprule
        Method & Identification Rate (\%) & SR (\%) \\ \midrule
        VLN-AVP (w/o STM) & 57.7 & 60 \\
        VLN-AVP & 84.6 & 85 \\ 
        \bottomrule
      \end{tabular}
      \begin{minipage}{0.9\linewidth}
            \small
            \vspace{1ex}
            `w/o STM' is without Short-Term Memory. 
       \end{minipage}
    \end{table}

    To further analyze how short-term memory (STM) augments the VLM's decision-making, we selected 20 critical road junctions and compared the VLM's identification rate of critical semantic cues and the success rate in making the right decision between our full method and the one without STM. The results indicate that STM significantly improves the identification rate of semantic visual cues, especially signs. By maintaining a cache of recent observations, STM ensures that the navigation cues and logic remain consistent even when direct visual evidence is temporarily missing in the current frame. 

    \section{Conclusion}
    In this paper, we introduced VLN-AVP, a zero-shot vision-language navigation framework designed for autonomous valet parking in unknown, map-free environments. By employing VLM as the decision core, we equipped AVP systems with semantic navigation capabilities that exceed previous approaches. Additionally, we proposed a hybrid long-short-term memory system that addresses the limitations of traditional single-frame decision-making and enables our system to learn from past experiences. We also constructed the largest underground garage dataset to date and the first VLN benchmark for underground parking. Extensive experiments in simulations and real-world settings demonstrated that our method outperforms existing methods in terms of feasibility and effectiveness. Future work will explore additional training to address the VLM's limitations in directional and spatial reasoning, and we plan to further improve and publicly release the VLN-AVP dataset. 


    \bibliographystyle{IEEEtran}
    \bibliography{references_short}
\end{document}